\documentclass[10pt, a4paper]{article}

\usepackage{lrec-coling2024} 
\usepackage{graphicx}
\usepackage{multirow}
\usepackage{amsmath}
\usepackage{amssymb}
\usepackage{subfig}
\usepackage{booktabs}

\title{Article Classification with Graph Neural Networks and Multigraphs}

\name{Khang Ly, Yury Kashnitsky, Savvas Chamezopoulos, Valeria Krzhizhanovskaya} 

\address{Elsevier B.V., Elsevier B.V., Elsevier B.V., Unversity of Amsterdam \\
         \{k.ly, y.kashnitskiy, s.chamezopoulos\}@elsevier.com, v.krzhizhanovskaya@uva.nl}

\abstract{
Classifying research output into context-specific label taxonomies is a challenging and relevant downstream task, given the volume of existing and newly published articles. We propose a method to enhance the performance of article classification by enriching simple Graph Neural Network (GNN) pipelines with multi-graph representations that simultaneously encode multiple signals of article relatedness, e.g. references, co-authorship, shared publication source, shared subject headings, as distinct edge types. Fully supervised transductive node classification experiments are conducted on the Open Graph Benchmark \texttt{OGBN-arXiv} dataset and the \texttt{PubMed} diabetes dataset, augmented with additional metadata from Microsoft Academic Graph and PubMed Central, respectively. The results demonstrate that multi-graphs consistently improve the performance of a variety of GNN models compared to the default graphs. When deployed with SOTA textual node embedding methods, the transformed multi-graphs enable simple and shallow 2-layer GNN pipelines to achieve results on par with more complex architectures.
 \\ \newline \Keywords{Heterogeneous Graph Learning, Graph Neural Networks, Article Classification, Document Relatedness} }

\begin{document}

\maketitleabstract

\section{Introduction}
Article classification is a challenging downstream task within natural language processing (NLP)~\citep{MIRONCZUK201836}. An important practical application is classifying existing or newly-published articles according to specific research taxonomies. The task can be approached as a graph node classification problem, where nodes represent articles with corresponding feature mappings, and edges are defined by a strong signal of article relatedness, e.g. citations/references. Conventionally, graph representation learning is handled in two phases: unsupervised node feature generation, followed by supervised learning on said features using the graph structure. Graph neural networks (GNNs) can be successfully employed for the second phase of such problems, being capable of preserving the rich structural information encoded by graphs. In recent years, prolific GNN architectures have achieved strong performance on citation network benchmarks~\citep{GCN, GraphSAGE, GAN, SIGN, RevGAT}. 

\begin{figure}[ht]
    \centering
    \includegraphics[width=0.45\textwidth]{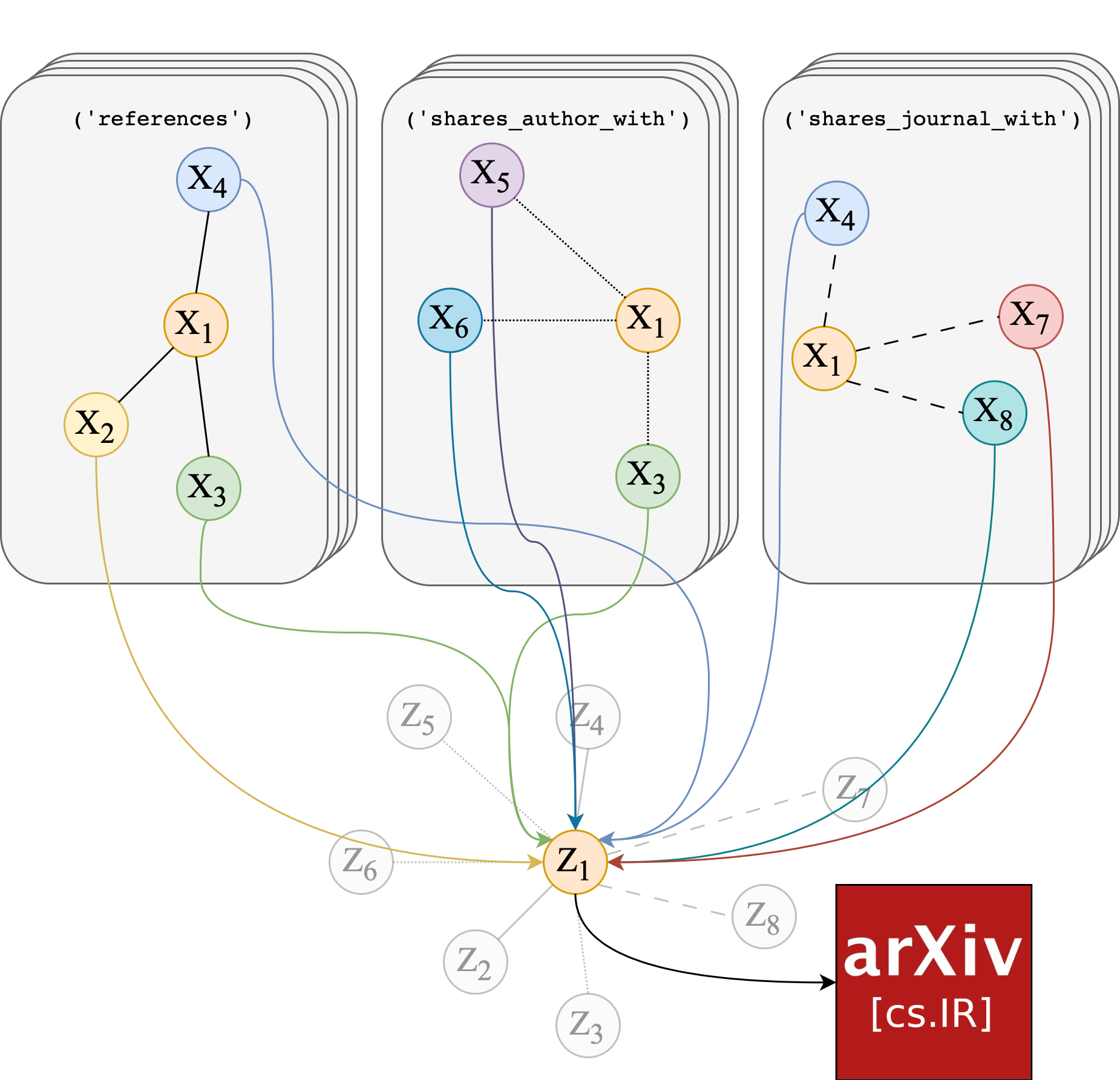}
    \caption{Illustration of the proposed multi-graph input, which enables the neighboring feature aggregation for a node $X_1$ to be performed across a variety of subgraphs, leveraging multiple signals of article relatedness (References, Authorship, and shared Journal depicted here).}
    \label{fig:methodology}
\end{figure}

We focus on combining textual information from articles with various indicators of article relatedness (citation data, co-authorship, subject fields, and publication sources) to create a graph with multiple edge types, also known as multi-graphs or heterogeneous graphs~\citep{Barabasi_Posfai_2017}. 
We use two established node classification benchmarks - the citation graphs \texttt{OGBN-arXiv} and \texttt{PubMed} - and leverage their connection to large citation databases - Microsoft Academic Graph (MAG) and PubMed Central - to retrieve the metadata fields and enrich the graph structure with additional edge types~\citep{OGB, Sen_Namata_Bilgic_Getoor_Galligher_Eliassi-Rad_2008}.
For node feature generation, we experiment with two approaches based on language model (LM) fine-tuning for graph representation learning - SimTG and TAPE - to infer embeddings based on articles' titles and abstracts, with the intention of capturing higher-order semantics compared to the defaults~\cite{duan2023simteg, he2024harnessing}. 
We test our transformed graphs with a variety of GNN backbone models, converted to support heterogeneous input using the relational graph convolutional network (R-GCN) framework~\citep{RGCN}. In essence, we approach a typically homogeneous task using heterogeneous techniques. The method is intuitively simple and interpretable; we do not utilize complex model architectures and training frameworks, focusing primarily on data retrieval and preprocessing to boost the performance of simpler models, thus maintaining a reasonably low computational cost and small number of fitted parameters.

A considerable volume of research is devoted to article classification, graph representation learning with respect to citation networks, and the adaptation of these practices to heterogeneous graphs~\citep{DBLP:journals/corr/abs-1901-00596, HGNNSurvey}. However, the application of \textit{heterogeneous} graph enrichment techniques to article classification is not well-studied and presents a research opportunity. Existing works on heterogeneous graphs often consider multiple node types, expanding from article to entity classification; we exclusively investigate the heterogeneity of paper-to-paper relationships to remain consistent with the single-node type problem setting. The emergence of rich metadata repositories for papers, e.g. OpenAlex, illustrates the relevance of our research~\citep{priem2022openalex}.

Scalability is often a concern with GNN architectures. For this reason, numerous approaches simplify typical GNN architectures with varying strategies, e.g. pre-computation or linearization, without sacrificing significant performance in most downstream tasks~\citep{SIGN, SGConv, prieto2023parameter}. Other solutions avoid GNNs altogether, opting for simpler approaches based on early graph-based techniques like label propagation, which outperform GNNs in several node classification benchmarks~\citep{CorrectSmooth}. The success of these simple approaches raises questions about the potential impracticality of deep GNN architectures on large real-world networks with a strong notion of locality, and whether or not such architectures are actually necessary to achieve satisfactory performance. 

Compared to simple homogeneous graphs, heterogeneous graphs encode rich structural and semantic information, and are more representative of real-world information networks and entity relationships~\citep{HGNNSurvey}. For example, networks constructed from citation databases can feature relations between papers, their authors, and shared keywords, often expressed in an RDF triple, e.g. ``\textit{paper} $\xrightarrow{\text{(co-)authored by}}$ \textit{author},'' ``\textit{paper} $\xrightarrow{\text{includes}}$ \textit{keyword},'' ``\textit{paper} $\xrightarrow{\text{cites}}$ \textit{paper}.'' Heterogeneous GNN architectures share many similarities with their homogeneous counterparts; a common approach is to aggregate feature information from local neighborhoods, while using additional modules to account for varying node and/or edge types~\citep{SeHGNN}. Notably, the relational graph convolutional network approach (R-GCN) by~\citet{RGCN} shows that GCN-based frameworks can be effectively applied to modeling relational data, specifically for the task of node classification. The authors propose a modeling technique where the message passing functions are duplicated and applied individually to each relationship type. This transformation can be generalized to a variety of GNN convolutional operators in order to convert them into their relational (heterogeneous) counterparts.

\section{Methodology}

We propose an approach focusing on dataset provenance, leveraging their linkage to large citation and metadata repositories, e.g. MAG and PubMed Central, to retrieve additional features and enrich their graph representations. The proposed method is GNN-agnostic, compatible with a variety of model pipelines (provided they can function with heterogeneous input) and textual node embedding techniques (results are presented with the provided features, plus the SimTG and TAPE embeddings). Figure~\ref{fig:methodology} provides a high-level overview of the method. 

The tested GNN backbones (see Section~\ref{sec:results}) are converted to support heterogeneous input using the aforementioned R-GCN transformation defined by~\citet{RGCN}, involving the duplication of the message passing functions at each convolutional layer per relationship type; we employ the PyTorch Geometric (PyG) implementation of this technique, using the mean as the aggregation operator~\citep{Fey:2019wv}.

\subsection{Datasets}
\begin{table*}[ht]
\centering
\resizebox{\textwidth}{!}{%
\begin{tabular}{@{}l|l|l|l|l|l|l|l@{}}
\toprule
Dataset                     & Edge Type & $|N|_{LCC}$ & $|E|_{LCC}$ & $|E|$      & Avg. Degree & Avg. Clust. Coeff. & Homophily \\ \midrule
\multirow{4}{*}{\texttt{OGBN-arXiv}} & References     & 169,343     & 2,315,598   & 2,315,598  & 13.7        & 0.310              & 0.654     \\ \cmidrule(l){2-8} 
                            & Authorship     & 145,973     & 6,697,998   & 6,749,335  & 39.9        & 0.775              & 0.580     \\ \cmidrule(l){2-8} 
                            & Source     & 63          & 3,906       & 605,660    & 3.6         & 1                  & 0.590     \\ \cmidrule(l){2-8} 
                            & Subject Area     & 144,529     & 8,279,492   & 8,279,687  & 48.9        & 0.630              & 0.319     \\ \midrule \midrule
\multirow{4}{*}{\texttt{PubMed}}     & References     & 19,716      & 88,649      & 88,649     & 4.5         & 0.246                   & 0.802     \\ \cmidrule(l){2-8} 
                            & Authorship     & 17,683      & 729,468     & 731,376    & 37.1        & 0.705                   & 0.721     \\ \cmidrule(l){2-8} 
                            & Source     & 2,213       & 4,895,156   & 11,426,930 & 579.6       & 1                  & 0.414     \\ \cmidrule(l){2-8} 
                            & Subject Area     & 18,345      & 1,578,526   & 1,578,530  & 80.1        & 0.481                   & 0.550     \\ \bottomrule
\end{tabular} }
\caption{\label{tab:metrics}Properties of constructed subgraphs: number of nodes and edges (post-conversion to undirected) in the largest connected component (LCC), total number of edges, average degree, network average clustering coefficient~\citep{clustering-coeff}, and edge homophily ratio (fraction of edges connecting nodes with the same label)~\citep{GCNHomophily}. Note that the References subgraphs are the only ones without isolated nodes. Note that the network average clustering coefficient computation \textit{excludes} isolated nodes with zero local clustering.}
\end{table*}

Our experiments are conducted on two datasets: the Open Graph Benchmark (OGB) \texttt{OGBN-arXiv} dataset and the \texttt{PubMed} diabetes dataset.

The OGB \texttt{OGBN-arXiv} dataset consists of 169,343 Computer Science papers from arXiv, hand-labeled into 40 subject areas by paper authors and arXiv moderators, with 1,166,243 reference links~\citep{OGB}. Default node features are constructed from textual information by averaging the embeddings of words (which are generated with the Skip-Gram model) in the articles' titles and abstracts. The dataset provides the mapping used between papers' node IDs and their original MAG IDs, which can be used to retrieve additional metadata.

The \texttt{PubMed} diabetes dataset consists of 19,717 papers from the National Library of Medicine's (NLM) PubMed database labeled into one of three categories: ``Diabetes Mellitus, Experimental,'' ``Diabetes Mellitus Type 1,'' and ``Diabetes Mellitus Type 2,'' with 44,338 references links~\citep{Sen_Namata_Bilgic_Getoor_Galligher_Eliassi-Rad_2008}. TF-IDF weighted word vectors from a dictionary of 500 unique words are provided as default node features. Similarly, the papers' original PubMed IDs can be used to fetch relevant metadata.

\subsection{Data Augmentation}\label{sec:data_aug}
For \texttt{OGBN-arXiv}, we used a July-2020 snapshot of the complete Microsoft Academic Graph (MAG) index (240M papers) - since MAG (and the associated API) was discontinued later - to obtain additional metadata~\citep{OAG}\footnote{The data is hosted by \href{https://www.aminer.cn/oag-2-1}{AMiner's \textbf{Open Academic Graph} project}. All chunks were downloaded locally and metadata of IDs corresponding to papers in \texttt{OGBN-arXiv} were saved.}. Potential indicators of paper relatedness include: authors, venue, and fields of study. Fields of study, e.g. ``computer science,'' ``neural networks,'' etc. are automatically assigned with an associated confidence score (which we do not use), and each paper can have multiple fields of study, making them functionally similar to keywords. Other metadata (DOI, volume, page numbers, etc.) are not useful for our purposes.

For \texttt{PubMed}, an unprocessed version of the dataset preserving the original paper IDs was used~\citep{namata:mlg12}\footnote{This version of the dataset is hosted by the \href{https://linqs.org/datasets/\#pubmed-diabetes}{LINQS Statistical Relational Learning Group}. The 2023 annual baseline on the NLM FTP server is accessed to retrieve metadata. All files were downloaded locally and metadata of matching IDs were extracted (19,716 records matched, 1 missing).}. A January-2023 snapshot of the complete PubMed citation database (35M papers) was accessed for additional metadata. Potential indicators of paper relatedness include: authors, journal (indicated by unique NLM journal IDs), and Medical Subject Headings (MeSH\textregistered). The latter is an NLM-controlled hierarchical vocabulary used to characterize biomedical article content.

Given the features of interest, we define three additional edge types for each dataset:

\begin{itemize}
    \item (Co)-Authorship: Two papers are connected if they share an author. This is based on the assumption that a given author tends to perform research on similar disciplines. Note that unlike MAG, PubMed Central does not provide unique identifiers for authors, so exact author names are used for \texttt{PubMed}, which can lead to some ambiguity in a minority of cases, e.g. two distinct authors with the same name. 
    \item Source: Two papers are connected if they were published at the same venue (\texttt{OGBN-arXiv}), or in the same journal (\texttt{PubMed}), with the intuition that specific conferences and journals feature papers contributing to similar research areas. 
    \item Subject Area: Two papers are connected if they share at least one field of study (\texttt{OGBN-arXiv}), or medical subject heading (\texttt{PubMed}).
\end{itemize}

\begin{figure*}[t]%
    \begin{center}
    \makebox[\textwidth][c]{
        \subfloat{{\includegraphics[width=0.5\textwidth]{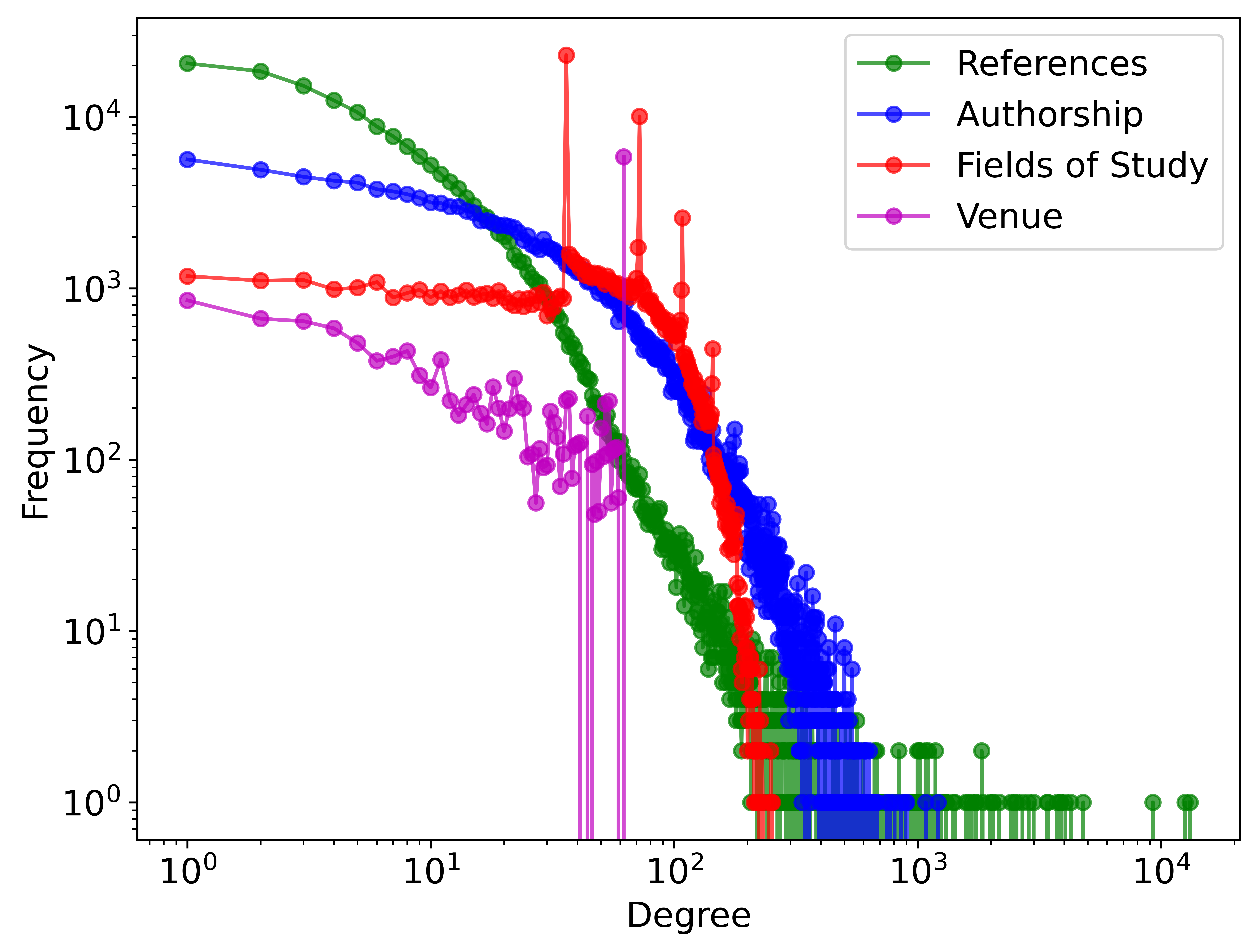} }}%
        \subfloat{{\includegraphics[width=0.5\textwidth]{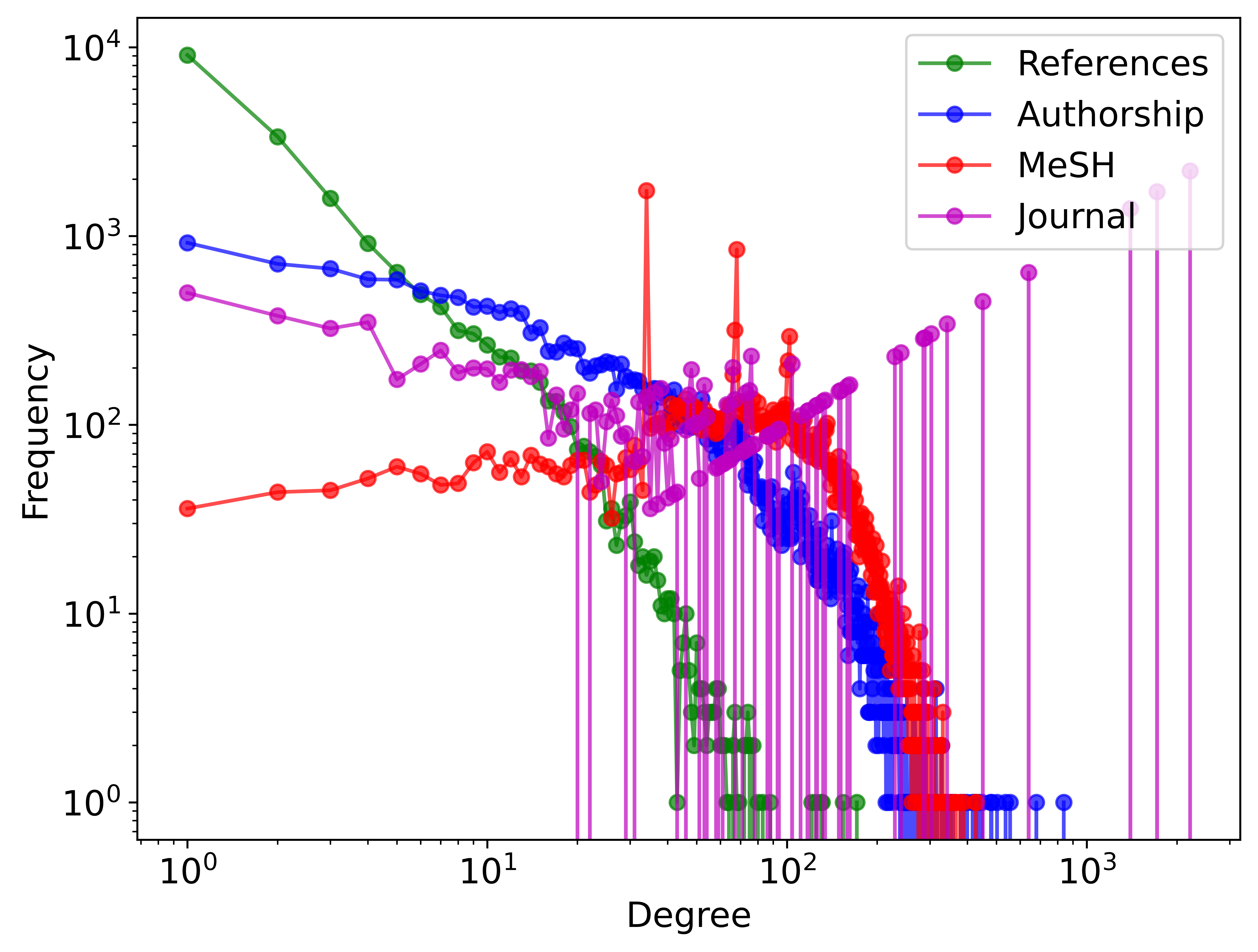} }}%
    }
    \caption{Degree distribution, i.e. frequency of each degree value, of all subgraphs for \texttt{OGBN-arXiv} (left) and \texttt{PubMed} (right), plotted on a log-log scale. Points indicate the unique degree values.}%
    \label{fig:degrees}%
    \end{center}
\end{figure*}

Since the \texttt{OGBN-arXiv} Source and both datasets' Subject Area relationships result in massive edge lists, posing out-of-memory issues on the utilized hardware, we only create edges between up to $k$ nodes per unique venue/field of study/MeSH, where $k$ is the mean number of papers per venue/field of study/MeSH, in order to reduce the subgraphs' sizes.

In a traditional citation network, the edges are typically directed, but in our experiments, they are undirected to strengthen the connections of communities in the graph. The graph includes only one node type, ``paper.'' Other approaches, notably in the citation recommendation domain, leverage node types to represent authors and journals~\citep{CoAuthorship}. However, this work strictly concerns relationships between papers and not between papers and other entities, in order to apply the homogeneous problem settings. Practically, the resultant graph would contain too many nodes, while the number of features and metadata is insufficient to generate informative representations of other node types, limiting their usefulness in the feature aggregation step. Hence, we specify our transformed graph as a multi-graph, i.e. possessing one node set, with distinct edges that are permitted to connect the same pair of nodes, and not a ``true'' heterogeneous graph.

For textual node feature representation, we leverage the recent SimTG and TAPE frameworks, which both utilize the raw textual features of datasets, in the form of concatenated titles and abstracts, and focus on fine-tuning pre-trained LMs and utilizing their last hidden states to infer node embeddings for training GNNs~\citep{duan2023simteg, he2024harnessing}. These methods are present in the top \texttt{OGBN-arXiv} leaderboard submissions (at the time of writing). SimTG performs supervised parameter-efficient fine-tuning (PEFT) of an LM on the article classification task. Pre-computed embeddings for \texttt{OGBN-arXiv} are provided (using \texttt{e5-large}), which we utilize here. Since the authors do not report results on \texttt{PubMed}, we reproduce their methods to generate embeddings for this dataset, using a SciBERT model~\citep{SciBERT}. TAPE proposes an LLM-to-LM interpreter, prompting GPT3.5 to perform zero-shot article classification and generate textual explanations for its decision-making process; the GPT3.5 predicted labels and explanations are then used to fine-tune a DeBERTa model. In this case, pre-computed embeddings were available for both datasets.

\subsection{Subgraph Properties}\label{sec:metrics}

\begin{table*}[t]
\centering
\resizebox{\textwidth}{!}{%
\begin{tabular}{@{}lllc|llc|llc@{}}
\toprule
\multicolumn{4}{c|}{Edge Types} &
  \multicolumn{3}{c|}{\texttt{OGBN-arXiv}} &
  \multicolumn{3}{c}{\texttt{PubMed}} \\ \midrule
\multicolumn{1}{c|}{Refs.} &
  \multicolumn{1}{c|}{Auth.} &
  \multicolumn{1}{c|}{Srce.} &
  Subj. &
  \multicolumn{1}{c|}{Default} &
  \multicolumn{1}{c|}{SimTG} &
  TAPE &
  \multicolumn{1}{c|}{Default} &
  \multicolumn{1}{c|}{SimTG} &
  TAPE \\ \midrule
\multicolumn{1}{c|}{\checkmark} &
  \multicolumn{1}{c|}{-} &
  \multicolumn{1}{c|}{-} &
  - &
  \multicolumn{1}{c|}{69.55 ± 0.31} &
  \multicolumn{1}{c|}{74.07 ± 0.03} &
  73.97 ± 0.01 &
  \multicolumn{1}{c|}{87.72 ± 0.29} &
  \multicolumn{1}{c|}{93.21 ± 0.24} &
  93.04 ± 0.01 \\ \midrule
\multicolumn{1}{c|}{-} &
  \multicolumn{1}{c|}{\checkmark} &
  \multicolumn{1}{c|}{-} &
  - &
  \multicolumn{1}{c|}{61.51 ± 0.15} &
  \multicolumn{1}{c|}{67.29 ± 0.23} &
  67.50 ± 0.09 &
  \multicolumn{1}{c|}{79.99 ± 0.29} &
  \multicolumn{1}{c|}{82.93 ± 0.09} &
  83.28 ± 0.03 \\ \midrule
\multicolumn{1}{c|}{-} &
  \multicolumn{1}{c|}{-} &
  \multicolumn{1}{c|}{\checkmark} &
  - &
  \multicolumn{1}{c|}{53.78 ± 0.26} &
  \multicolumn{1}{c|}{73.30 ± 0.08} &
  73.05 ± 0.02 &
  \multicolumn{1}{c|}{48.99 ± 6.82*} &
  \multicolumn{1}{c|}{61.97 ± 0.44} &
  62.23 ± 1.37 \\ \midrule
\multicolumn{1}{c|}{-} &
  \multicolumn{1}{c|}{-} &
  \multicolumn{1}{c|}{-} &
  \checkmark &
  \multicolumn{1}{c|}{49.87 ± 0.11} &
  \multicolumn{1}{c|}{55.54 ± 0.31} &
  56.07 ± 0.32 &
  \multicolumn{1}{c|}{73.57 ± 0.42} &
  \multicolumn{1}{c|}{74.96 ± 1.90} &
  75.78 ± 0.12 \\ \midrule \midrule
\multicolumn{1}{c|}{\checkmark} &
  \multicolumn{1}{c|}{\checkmark} &
  \multicolumn{1}{c|}{-} &
  - &
  \multicolumn{1}{c|}{71.40 ± 0.19} &
  \multicolumn{1}{c|}{75.80 ± 0.11} &
  75.98 ± 0.06 &
  \multicolumn{1}{c|}{88.91 ± 0.18} &
  \multicolumn{1}{c|}{\textbf{93.62 ± 0.01}} &
  \textbf{93.45 ± 0.17} \\ \midrule
\multicolumn{1}{c|}{\checkmark} &
  \multicolumn{1}{c|}{-} &
  \multicolumn{1}{c|}{\checkmark} &
  - &
  \multicolumn{1}{c|}{68.72 ± 0.30} &
  \multicolumn{1}{c|}{76.05 ± 0.02} &
  75.85 ± 0.10 &
  \multicolumn{1}{c|}{87.97 ± 0.25} &
  \multicolumn{1}{c|}{93.14 ± 0.05} &
  93.05 ± 0.20 \\ \midrule
\multicolumn{1}{c|}{\checkmark} &
  \multicolumn{1}{c|}{-} &
  \multicolumn{1}{c|}{-} &
  \checkmark &
  \multicolumn{1}{c|}{70.01 ± 0.05} &
  \multicolumn{1}{c|}{74.42 ± 0.08} &
  74.45 ± 0.05 &
  \multicolumn{1}{c|}{88.14 ± 0.06} &
  \multicolumn{1}{c|}{92.79 ± 0.29} &
  93.30 ± 0.11 \\ \midrule
\multicolumn{1}{c|}{\checkmark} &
  \multicolumn{1}{c|}{\checkmark} &
  \multicolumn{1}{c|}{\checkmark} &
  - &
  \multicolumn{1}{c|}{70.97 ± 0.19} &
  \multicolumn{1}{c|}{\textbf{77.26 ± 0.04}} &
  \textbf{77.14 ± 0.10} &
  \multicolumn{1}{c|}{88.86 ± 0.10} &
  \multicolumn{1}{c|}{93.54 ± 0.10} &
  92.89 ± 0.24 \\ \midrule
\multicolumn{1}{c|}{\checkmark} &
  \multicolumn{1}{c|}{\checkmark} &
  \multicolumn{1}{c|}{-} &
  \checkmark &
  \multicolumn{1}{c|}{\textbf{71.81 ± 0.06}} &
  \multicolumn{1}{c|}{75.94 ± 0.07} &
  76.15 ± 0.12 &
  \multicolumn{1}{c|}{\textbf{89.22 ± 0.15}} &
  \multicolumn{1}{c|}{93.11 ± 0.32} &
  93.23 ± 0.15 \\ \midrule
\multicolumn{1}{c|}{\checkmark} &
  \multicolumn{1}{c|}{-} &
  \multicolumn{1}{c|}{\checkmark} &
  \checkmark &
  \multicolumn{1}{c|}{69.30 ± 0.14} &
  \multicolumn{1}{c|}{76.08 ± 0.05} &
  75.97 ± 0.05 &
  \multicolumn{1}{c|}{88.35 ± 0.17} &
  \multicolumn{1}{c|}{92.84 ± 0.16} &
  92.73 ± 0.50 \\ \midrule
\multicolumn{1}{c|}{\checkmark} &
  \multicolumn{1}{c|}{\checkmark} &
  \multicolumn{1}{c|}{\checkmark} &
  \checkmark &
  \multicolumn{1}{c|}{71.30 ± 0.08} &
  \multicolumn{1}{c|}{77.17 ± 0.02} &
  77.07 ± 0.11 &
  \multicolumn{1}{c|}{88.47 ± 0.28} &
  \multicolumn{1}{c|}{93.16 ± 0.29} &
  93.12 ± 0.16 \\ \bottomrule
\end{tabular} }
\caption{\label{tab:ablation} References, Authorship, Source (venue or journal), and Subject Area (fields of study or MeSH) subgraph ablation study for both datasets, 3-run average test accuracy with a 2-layer GCN and consistent hyperparameter values per dataset. The best results for each column are highlighted in bold. Asterisk indicates (significant) overfitting and instability.}
\end{table*}

Some insights on the characteristics of the defined subgraphs can be derived from Table~\ref{tab:metrics}. While the References graphs do not exhibit the tight clustering typical of real-world information networks, the strong signal of relatedness in the edges has nonetheless ensured their compatibility with message passing GNN paradigms~\citep{DBLP:journals/corr/abs-1901-00596}. This relatedness is also evident in the Authorship graphs, and the high level of clustering confirms the initial hypothesis that researchers co-author papers within similar topics. The Subject Area relationships, include many edges formed between shared generic keywords, e.g. ``computer science,'' leading to rather average homophily. The Source subgraphs consist of isolated fully-connected clusters per unique source, with no inter-cluster connections, as each paper belongs to only one journal or venue. As with the Subject Area relationships, the research scope covered by a given publication conference or journal can be quite broad with respect to the paper labels.

Figure~\ref{fig:degrees} shows the degree distribution of all edge type subgraphs in both datasets, which gives a clear view of the subgraphs' structures when interpreted with the above metrics. The high frequency of large node degrees in the \texttt{PubMed} Source subgraph corresponds to large journals; the size of the LCC (2,213) is the number of papers in the largest journal. While not visible for the \texttt{OGBN-arXiv} Source subgraph due to the aforementioned sampling in Section~\ref{sec:data_aug}, a similar distribution would occur for large venues if all possible edges had been included. In contrast, the lower occurrence of high degree nodes and low clustering in the References subgraphs of both datasets indicates greater average distance across the LCC compared to the other subgraphs; such a structure stands to benefit the most from the multi-hop neighborhood feature aggregation performed by GNNs. Relative to the References, the Authorship and Subject Area subgraphs exhibit increased skewness in the distribution and higher average clustering, which indicates the presence of more (near-)cliques, i.e. subsections of the graph wherein (almost) any two papers share an author or topic. Hence, these subgraphs bear the closest structural resemblance to small-world networks~\citep{watts_collective_1998}. The impact of these degree distributions on classification performance is further investigated in Section~\ref{sec:ablation}.

\section{Experiments and Results}\label{sec:results}

We evaluate model performance on the task of \textit{fully supervised transductive node classification}. The metric is multi-class accuracy on the test set. The proposed data preparation scheme is tested with several GNN architectures commonly deployed in benchmarks. We consider two GCN setups (base one and with a jumping knowledge module using max-pooling as the aggregation scheme), as well as GraphSAGE~\citep{GCN, JK, GraphSAGE}. We also run experiments with the simplified graph convolutional operator (SGC)~\citep{SGConv}. The increased graph footprint can lead to scalability concerns, hence the performance of such lightweight and parameter-efficient methods is of interest.

For \texttt{OGBN-arXiv}, the provided time-based split is used: train on papers published until 2017, validate on those published in 2018, test on those published since 2019. For \texttt{PubMed}, nodes of each class are randomly split into 60\% - 20\% - 20\% for training - validation - and testing. Ablation experiments are also performed to examine the impact of the different edge types (averaged across 3 runs) and to identify the optimal edge type configuration for both datasets, on which we then report final results (averaged across 10 runs). Experiments were conducted on a \texttt{g4dn.2xlarge} EC2 instance (32~GB RAM, 1 NVIDIA Tesla T4 16~GB VRAM). Models are trained with negative log-likelihood loss, early stopping based on validation accuracy (patience of 20 epochs, with an upper limit of 500 epochs), and linear learning rate scheduling.

\subsection{Ablation Study}\label{sec:ablation}

\begin{table*}[t]
\centering
\resizebox{\textwidth}{!}{%
\begin{tabular}{@{}c|c|llc|lllllc@{}}
\toprule
\multirow{2}{*}{Dataset} &
  \multirow{2}{*}{GNN} &
  \multicolumn{3}{c|}{Default Graph} &
  \multicolumn{6}{c}{Multi-graph} \\ \cmidrule(l){3-11} 
 &
   &
  \multicolumn{1}{c|}{Default} &
  \multicolumn{1}{c|}{SimTG} &
  TAPE &
  \multicolumn{1}{c|}{Default} &
  \multicolumn{1}{c|}{$\Delta$} &
  \multicolumn{1}{c|}{SimTG} &
  \multicolumn{1}{c|}{$\Delta$} &
  \multicolumn{1}{c|}{TAPE} &
  $\Delta$ \\ \midrule
\multirow{4}{*}{\texttt{OGBN-arXiv}} &
  GCN &
  \multicolumn{1}{c|}{69.67 ± 0.17} &
  \multicolumn{1}{c|}{73.98 ± 0.11} &
  74.08 ± 0.10 &
  \multicolumn{1}{c|}{71.88 ± 0.06} &
  \multicolumn{1}{c|}{\colorbox{green}{+2.21\%}} &
  \multicolumn{1}{c|}{77.30 ± 0.09} &
  \multicolumn{1}{c|}{\colorbox{green}{+3.32\%}} &
  \multicolumn{1}{c|}{77.10 ± 0.10} &
  \colorbox{green}{+3.02\%} \\ \cmidrule(l){2-11} 
 &
  GCN+JK &
  \multicolumn{1}{c|}{70.24 ± 0.17} &
  \multicolumn{1}{c|}{75.01 ± 0.15} &
  75.15 ± 0.16 &
  \multicolumn{1}{c|}{71.56 ± 0.21} &
  \multicolumn{1}{c|}{\colorbox{green}{+1.32\%}} &
  \multicolumn{1}{c|}{77.05 ± 0.10} &
  \multicolumn{1}{c|}{\colorbox{green}{+2.04\%}} &
  \multicolumn{1}{c|}{76.66 ± 0.10} &
  \colorbox{green}{+1.51\%} \\ \cmidrule(l){2-11} 
 &
  SAGE &
  \multicolumn{1}{c|}{68.99 ± 0.18} &
  \multicolumn{1}{c|}{75.65 ± 0.11} &
  75.41 ± 0.13 &
  \multicolumn{1}{c|}{71.37 ± 0.21} &
  \multicolumn{1}{c|}{\colorbox{green}{+2.38\%}} &
  \multicolumn{1}{c|}{77.39 ± 0.15} &
  \multicolumn{1}{c|}{\colorbox{green}{+1.74\%}} &
  \multicolumn{1}{c|}{76.68 ± 0.06} &
  \colorbox{green}{+1.27\%} \\ \cmidrule(l){2-11} 
 &
  SGC &
  \multicolumn{1}{c|}{68.73 ± 0.14} &
  \multicolumn{1}{c|}{73.95 ± 0.03} &
  73.65 ± 0.25 &
  \multicolumn{1}{c|}{70.24 ± 0.05} &
  \multicolumn{1}{c|}{\colorbox{green}{+1.51\%}} &
  \multicolumn{1}{c|}{77.24 ± 0.01} &
  \multicolumn{1}{c|}{\colorbox{green}{+3.29\%}} &
  \multicolumn{1}{c|}{75.93 ± 0.17} &
  \colorbox{green}{+2.28\%} \\ \midrule \midrule
\multirow{4}{*}{\texttt{PubMed}} &
  GCN &
  \multicolumn{1}{c|}{87.67 ± 0.25} &
  \multicolumn{1}{c|}{92.92 ± 0.12} &
  92.92 ± 0.17 &
  \multicolumn{1}{c|}{89.15 ± 0.14} &
  \multicolumn{1}{c|}{\colorbox{green}{+1.48\%}} &
  \multicolumn{1}{c|}{93.49 ± 0.16} &
  \multicolumn{1}{c|}{\colorbox{green}{+0.57\%}} &
  \multicolumn{1}{c|}{93.59 ± 0.26} &
  \colorbox{green}{+0.67\%} \\ \cmidrule(l){2-11} 
 &
  GCN+JK &
  \multicolumn{1}{c|}{87.13 ± 0.28} &
  \multicolumn{1}{c|}{93.68 ± 0.18} &
  93.49 ± 0.36 &
  \multicolumn{1}{c|}{87.53 ± 0.62} &
  \multicolumn{1}{c|}{\colorbox{lightgray}{+0.40\%}} &
  \multicolumn{1}{c|}{94.11 ± 0.18} &
  \multicolumn{1}{c|}{\colorbox{green}{+0.43\%}} &
  \multicolumn{1}{c|}{94.17 ± 0.13} &
  \colorbox{green}{+0.68\%} \\ \cmidrule(l){2-11} 
 &
  SAGE &
  \multicolumn{1}{c|}{88.30 ± 0.10} &
  \multicolumn{1}{c|}{95.46 ± 0.07} &
  94.87 ± 0.10 &
  \multicolumn{1}{c|}{89.75 ± 0.09} &
  \multicolumn{1}{c|}{\colorbox{green}{+1.45\%}} &
  \multicolumn{1}{c|}{95.51 ± 0.10} &
  \multicolumn{1}{c|}{\colorbox{lightgray}{+0.05\%}} &
  \multicolumn{1}{c|}{94.93 ± 0.13} &
  \colorbox{lightgray}{+0.06\%} \\ \cmidrule(l){2-11} 
 &
  SGC &
  \multicolumn{1}{c|}{86.87 ± 0.16} &
  \multicolumn{1}{c|}{90.31 ± 0.30} &
  90.57 ± 0.32 &
  \multicolumn{1}{c|}{86.56 ± 0.57} &
  \multicolumn{1}{c|}{\colorbox{red}{-0.31\%}} &
  \multicolumn{1}{c|}{91.41 ± 0.13} &
  \multicolumn{1}{c|}{\colorbox{green}{+1.10\%}} &
  \multicolumn{1}{c|}{91.20 ± 0.21} &
  \colorbox{green}{+0.63\%} \\ \bottomrule
\end{tabular} }
\caption{\label{tab:gnn_results} Results with a variety of GNN backbones on the best multi-graph configuration per embedding method, based on the ablation study in Table~\ref{tab:ablation}, so e.g. the multi-graph for \texttt{OGBN-arXiv} GCN with SimTG embeddings consists of the References, Authorship, and Source graphs; the same multi-graph configuration is re-used for all other GNNs trained on \texttt{OGBN-arXiv} with SimTG embeddings. The baseline results on the default graph and the accuracy difference over the baseline are also displayed per embedding method. Green, gray, and red indicate increase, insignificant increase, and decrease, respectively.}
\end{table*}

Ablation results for both datasets are presented in Table~\ref{tab:ablation}, separated by node embedding method. First, all possible homogeneous subgraphs are inspected, as this is the conventional input data for this task (see the first 4 rows). The best performance is consistently achieved on the References graphs. Then we build upon the References graph by adding different combinations of other subgraphs. The results demonstrate that transitioning to multi-graphs can yield up to 3.19\% performance improvement on \texttt{OGBN-arXiv} and 1.50\% on \texttt{PubMed} (see differences between References-only and bold configurations). These results were obtained with a 2-layer GCN base, using an initial learning rate of 0.001 and hidden feature dimensionality of 128. For \texttt{PubMed}, we add an optimizer weight decay of 0.005.

Cross-checking with the metrics in Table~\ref{tab:metrics} implies improvements from multi-graphs roughly correspond to the edge homophily ratio of the utilized subgraphs, as strong homophily is implicitly assumed by the neighborhood aggregation mechanism of GCN-based models. Subsequently, their performance can be erratic and unpredictable in graphs with comparatively low homophily~\citep{GCN, GCNHomophily}. Since the R-GCN transformation collects neighborhoods from input subgraphs with equal weighting, including a comparatively noisy subgraph, e.g. \texttt{PubMed} Source, can worsen predictive performance. Changing the R-GCN aggregation operator, e.g. from mean to concatenation, does not alleviate this.

The Source subgraphs benefit substantially from the LM-based features, as the extent of feature aggregation is comparatively limited, due to the aforementioned tight clustering and isolation. Hence, the classifier relies more on the raw separability of the textual node features. This also explains the breakdown in performance when using the \texttt{PubMed} Source subgraph in a homogeneous setting, as a paper might possess only a few non-zero feature dimensions when using the default word vectors. The Subject Area subgraphs are more structurally preferable, but noisy edges (from keywords tied to concepts that are higher-level than the paper labels) reduce their usefulness in classification. In addition, the semantic information they encode is dominated by the dense LM-based features, reflected by the fact that they only appear in the optimal multi-graph configuration when using the default embeddings. Across all experimental settings, the Authorship subgraph enables consistent gains, and can outperform configurations that use more subgraphs. These trends are expected, given the characteristics discussed in Section~\ref{sec:metrics}.

\subsection{Optimal Configuration}

Results with the optimal configuration identified from the ablation study are listed in Table~\ref{tab:gnn_results}, for both datasets.

In most cases, preliminary experiments indicated that deeper (3 or more layers) networks either worsen or do not benefit performance of the tested models in multi-graph configurations (however, note that tested single-layer models underfit and thus do not improve performance). Likely, the additional feature averaging step from the R-GCN transformation increases the risk of oversmoothing even on shallow networks. These hypothesized effects are more pronounced when using graphs with high average degree, e.g. the Source and Subject Area subgraphs; nodes with high degree aggregate more information from their neighbors, increasing the likelihood of homogenization as network depth increases~\citep{pmlr-v119-chen20v}.

The results demonstrate that the additional structural information provided by multi-graphs generally improves final performance of a variety of hetero-transformed GNN frameworks compared to their homogeneous counterparts on both datasets, with more pronounced effects on \texttt{OGBN-arXiv}, when making optimal subgraph choices (though, suboptimal choices can still situationally improve performance). These improvements are independent of the tested textual embedding methods, and can occur even when the added subgraphs possess suboptimal graph properties, e.g. lower edge homophily ratio and presence of isolated nodes, compared to the starting References graph. Notably, the best results are competitive with the SOTA, while operating on a limited compute budget and low level of complexity (simple 2-layer GNN model pipelines with comparatively few trainable parameters). On \texttt{OGBN-arXiv}, we can achieve a top-5 result (at the time of writing) deploying our multi-graph with a GraphSAGE backbone and SimTG embeddings. 

\section{Conclusions and Future Work}
In this paper, we propose a data transformation methodology leveraging metadata retrieved from citation databases to create enriched multi-graph representations based on various additional signals of document relatedness: co-authorship, publication source, fields of study, and subject headings. We also test the substitution of default node features with LM-based embeddings to capture higher-dimensionality textual semantics. By nature, the methodology is GNN- and embedding-agnostic. Deploying optimal configurations of the transformed multi-graph with a variety of simple GNN pipelines leads to consistent improvements over the starting graph, and enables results on par with the SOTA in full-supervised node classification. Overall, results show that our methodology can be an effective strategy to achieve respectable performance on datasets with readily-available article metadata, without necessitating complex GNN architectures and lengthy (pre-)training procedures.

As the methodology is compatible with any hetero-transformable GNN backbone and textual node embedding technique, we expect that deploying the transformed data with SOTA GNN frameworks, e.g. RevGAT by~\citet{RevGAT} on \texttt{OGBN-arXiv}, will lead to greater raw performance. Though, the larger memory footprint of the graph may complicate the application of such frameworks. 

Refining the edge type definitions, e.g. connect papers that share at least two fields of study and/or remove ``generic'' fields applicable to a majority of papers in the set, can help de-noising and improving the properties of the respective subgraphs. A custom aggregation scheme could be implemented for the heterogeneous transformation dependent on individual subgraph properties, such as a weighted average based on some metric of subgraph ``quality,'' e.g. homophily. To mitigate the increased risk of oversmoothing induced by multi-graphs and stabilize convergence behavior, additional regularization techniques, e.g. DropEdge by~\citet{Rong2020DropEdge:}, could be considered.

\nocite{*}
\section{References}\label{sec:reference}
\bibliographystyle{lrec-coling2024-natbib}
\bibliography{lrec-coling2024-example}

\section*{A. Reproducibility Statement}
For reproducibility, our implementation is available at \href{https://github.com/lyvykhang/edgehetero-nodeproppred}{this GitHub repository}, including the required external metadata, node feature embeddings, and other misc. information.

\section*{B. Supplementary Results}\label{apx:b}
The validation accuracy and number of trainable GNN parameters for all results in Table~\ref{tab:gnn_results} can be seen in Table~\ref{tab:val_accs}. Note that the feature dimensionality for SimTG differs between datasets - 1024-dim. (\texttt{e5-large}) for \texttt{OGBN-arXiv}, 768-dim. (\texttt{scibert-scivocab-uncased}) for \texttt{PubMed}.

\begin{table*}[h]
\resizebox{\textwidth}{!}{%
\begin{tabular}{@{}c|c|c|llc|llc@{}}
\toprule
\multirow{2}{*}{Dataset}    & \multirow{2}{*}{GNN}    & \multirow{2}{*}{Metric} & \multicolumn{3}{c|}{Default Graph}                                                   & \multicolumn{3}{c}{Multi-graph}                                                       \\ \cmidrule(l){4-9} 
                            &                         &                         & \multicolumn{1}{c|}{Default}    & \multicolumn{1}{c|}{SimTG}      & TAPE       & \multicolumn{1}{c|}{Default}    & \multicolumn{1}{c|}{SimTG}      & TAPE       \\ \midrule
\multirow{8}{*}{\texttt{OGBN-arXiv}} & \multirow{2}{*}{GCN}    & Val. Acc.               & \multicolumn{1}{c|}{70.78 ± 0.15} & \multicolumn{1}{c|}{75.51 ± 0.15} & 75.41 ± 0.06 & \multicolumn{1}{c|}{73.15 ± 0.12} & \multicolumn{1}{c|}{78.39 ± 0.16} & 78.07 ± 0.08 \\ \cmidrule(l){3-9} 
                            &                         & \# Params               & \multicolumn{1}{c|}{21,928}       & \multicolumn{1}{c|}{136,616}      & 103,848      & \multicolumn{1}{c|}{65,272}       & \multicolumn{1}{c|}{409,336}      & 311,032      \\ \cmidrule(l){2-9} 
                            & \multirow{2}{*}{GCN+JK} & Val. Acc.               & \multicolumn{1}{c|}{71.29 ± 0.09} & \multicolumn{1}{c|}{76.20 ± 0.12} & 76.10 ± 0.11 & \multicolumn{1}{c|}{73.13 ± 0.13} & \multicolumn{1}{c|}{78.38 ± 0.12} & 77.78 ± 0.07 \\ \cmidrule(l){3-9} 
                            &                         & \# Params               & \multicolumn{1}{c|}{38,686}       & \multicolumn{1}{c|}{153,384}      & 120,616      & \multicolumn{1}{c|}{104,744}      & \multicolumn{1}{c|}{448,808}      & 350,504      \\ \cmidrule(l){2-9} 
                            & \multirow{2}{*}{SAGE}   & Val. Acc.               & \multicolumn{1}{c|}{70.16 ± 0.25} & \multicolumn{1}{c|}{77.01 ± 0.15} & 76.27 ± 0.07 & \multicolumn{1}{c|}{72.90 ± 0.10} & \multicolumn{1}{c|}{78.69 ± 0.15} & 77.63 ± 0.04 \\ \cmidrule(l){3-9} 
                            &                         & \# Params               & \multicolumn{1}{c|}{43,432}       & \multicolumn{1}{c|}{272,808}      & 207,272      & \multicolumn{1}{c|}{129,784}      & \multicolumn{1}{c|}{817,912}      & 621,304      \\ \cmidrule(l){2-9} 
                            & \multirow{2}{*}{SGC}    & Val. Acc.               & \multicolumn{1}{c|}{69.82 ± 0.12} & \multicolumn{1}{c|}{75.04 ± 0.00} & 74.80 ± 0.16 & \multicolumn{1}{c|}{71.59 ± 0.06} & \multicolumn{1}{c|}{78.39 ± 0.00} & 77.11 ± 0.09 \\ \cmidrule(l){3-9} 
                            &                         & \# Params               & \multicolumn{1}{c|}{5,160}        & \multicolumn{1}{c|}{41,000}       & 30,760       & \multicolumn{1}{c|}{15,480}       & \multicolumn{1}{c|}{123,000}      & 92,280       \\ \midrule \midrule
\multirow{8}{*}{\texttt{PubMed}}     & \multirow{2}{*}{GCN}    & Val. Acc.               & \multicolumn{1}{c|}{88.17 ± 0.16} & \multicolumn{1}{c|}{93.67 ± 0.28} & 93.81 ± 0.09 & \multicolumn{1}{c|}{89.39 ± 0.16} & \multicolumn{1}{c|}{94.52 ± 0.06} & 94.63 ± 0.12 \\ \cmidrule(l){3-9} 
                            &                         & \# Params               & \multicolumn{1}{c|}{64,771}       & \multicolumn{1}{c|}{99,075}       & 99,075       & \multicolumn{1}{c|}{193,801}      & \multicolumn{1}{c|}{197,894}      & 197,894      \\ \cmidrule(l){2-9} 
                            & \multirow{2}{*}{GCN+JK} & Val. Acc.               & \multicolumn{1}{c|}{87.93 ± 0.18} & \multicolumn{1}{c|}{94.77 ± 0.15} & 94.56 ± 0.27 & \multicolumn{1}{c|}{88.74 ± 0.12} & \multicolumn{1}{c|}{95.30 ± 0.23} & 95.48 ± 0.21 \\ \cmidrule(l){3-9} 
                            &                         & \# Params               & \multicolumn{1}{c|}{81,539}       & \multicolumn{1}{c|}{115,843}      & 115,843      & \multicolumn{1}{c|}{242,819}      & \multicolumn{1}{c|}{230,787}      & 230,787      \\ \cmidrule(l){2-9} 
                            & \multirow{2}{*}{SAGE}   & Val. Acc.               & \multicolumn{1}{c|}{89.15 ± 0.24} & \multicolumn{1}{c|}{96.36 ± 0.17} & 96.27 ± 0.16 & \multicolumn{1}{c|}{90.64 ± 0.13} & \multicolumn{1}{c|}{96.57 ± 0.13} & 96.59 ± 0.05 \\ \cmidrule(l){3-9} 
                            &                         & \# Params               & \multicolumn{1}{c|}{129,155}      & \multicolumn{1}{c|}{197,763}      & 197,763      & \multicolumn{1}{c|}{386,953}      & \multicolumn{1}{c|}{395,270}      & 395,270      \\ \cmidrule(l){2-9} 
                            & \multirow{2}{*}{SGC}    & Val. Acc.               & \multicolumn{1}{c|}{87.05 ± 0.12} & \multicolumn{1}{c|}{91.00 ± 0.17} & 91.59 ± 0.17 & \multicolumn{1}{c|}{86.72 ± 0.39} & \multicolumn{1}{c|}{92.31 ± 0.06} & 92.49 ± 0.17 \\ \cmidrule(l){3-9} 
                            &                         & \# Params               & \multicolumn{1}{c|}{1,503}        & \multicolumn{1}{c|}{2,307}        & 2,307        & \multicolumn{1}{c|}{4,509}        & \multicolumn{1}{c|}{4,614}        & 4,614        \\ \bottomrule
\end{tabular}%
}
\caption{\label{tab:val_accs}Validation accuracy and number of trainable GNN parameters for all results in Table~\ref{tab:gnn_results}.}
\end{table*}

\begin{table*}[h]
\centering
\resizebox{0.6\textwidth}{!}{%
\begin{tabular}{@{}l|l|l|l|l|l@{}}
\toprule
Dataset                     & Hyperparameter      & GCN   & GCN+JK & SAGE  & SGC \\ \midrule
\multirow{5}{*}{\texttt{OGBN-arXiv}} & \# Layers           & 2     & 2      & 2     & 2   \\ \cmidrule(l){2-6} 
                            & Hidden Channels     & 128   & 128    & 128   & -   \\ \cmidrule(l){2-6} 
                            & Dropout             & 0     & 0      & 0     & -   \\ \cmidrule(l){2-6} 
                            & Init. Learning Rate & 0.001 & 0.001  & 0.001 & 0.1\textsuperscript{\textbf{*}} \\ \cmidrule(l){2-6} 
                            & Weight Decay        & 0     & 0      & 0     & 0   \\ \midrule \midrule
\multirow{5}{*}{\texttt{PubMed}}     & \# Layers           & 2     & 2      & 2     & 2   \\ \cmidrule(l){2-6} 
                            & Hidden Channels     & 128   & 128    & 128   & -   \\ \cmidrule(l){2-6} 
                            & Dropout             & 0     & 0.2\textsuperscript{\textbf{**}}    & 0     & -   \\ \cmidrule(l){2-6} 
                            & Init. Learning Rate & 0.001 & 0.001  & 0.001 & 0.1 \\ \cmidrule(l){2-6} 
                            & Weight Decay        & 0.005 & 0\textsuperscript{\textbf{***}}      & 0.005 & 0   \\ \bottomrule
\end{tabular}%
}
\caption{\label{tab:params}Hyperparameters used for all results in Table~\ref{tab:gnn_results}. \textsuperscript{\textbf{*}}0.01 for TAPE embeddings on multi-graph. \textsuperscript{\textbf{**}}0.5 for default embeddings on multi-graph. \textsuperscript{\textbf{***}}0.001 for default embeddings on multi-graph.}
\end{table*}

\section*{C. Hyperparameters}\label{apx:c}
The hyperparameters used for all results in Table~\ref{tab:gnn_results} can be seen in Table~\ref{tab:params}. Note that these parameters were not comprehensively tuned per (dataset, GNN, graph type) combination, to illustrate the generality of our methods; a more extensive hyperparameter search can yield better results. Also, expanding the ablation study in Table~\ref{tab:ablation} for specific GNN backbones rather than generalizing the GCN-applicable optimal configuration can further optimize results.

\end{document}